\documentclass[11pt,letterpaper]{article}
\usepackage{emnlp2016}
\usepackage{times}
\usepackage{latexsym}
\usepackage{graphics}
\usepackage{graphicx}
\usepackage[utf8]{inputenc}
\usepackage[table]{xcolor}
\usepackage{multirow}
\usepackage{slashbox}
\usepackage{float}
\emnlpfinalcopy
\usepackage{url}
\def\emnlppaperid{***}

\title{A Correlational Encoder Decoder Architecture for Pivot Based Sequence Generation}

 \author{Amrita Saha \\ IBM Research India \\ {\scriptsize{\tt amrsaha4@in.ibm.com}}
         \And  
 Mitesh M Khapra \\ IBM Research India \\ {\scriptsize{\tt mikhapra@in.ibm.com}}
 \And
 Sarath Chandar \\ University of Montreal \\ {\scriptsize{\tt apsarathchandar@gmail.com}}
 \AND
 Janarthanan Rajendran \\ Indian Institute of Technology Madras \\ {\scriptsize{\tt rsdjjana@gmail.com}}
\And
 Kyunghyun Cho \\ New York University\\ {\scriptsize{\tt kyunghyun.cho@nyu.edu}}
  }

\date{}

\begin{document}

\maketitle

\begin{abstract}
Interlingua based Machine Translation (MT) aims to encode multiple languages into a common linguistic representation and then decode sentences in multiple target languages from this representation.
In this work we explore this idea in the context of neural encoder decoder architectures, albeit on a smaller scale and without MT as the end goal. Specifically, we consider the case of three languages or modalities $X$, $Z$ and $Y$ wherein we are interested in generating sequences in $Y$ starting from information available in $X$. However, there is no parallel training data available between $X$ and $Y$ but, training data is available between $X$ \& $Z$ and $Z$ \& $Y$ (as is often the case in many real world applications). $Z$ thus acts as a pivot/bridge. An obvious solution, which is perhaps less elegant but works very well in practice is to train a two stage model which first converts from $X$ to $Z$ and then from $Z$ to $Y$. Instead we explore an interlingua inspired solution which \textbf{jointly learns} to do the following (i) encode $X$ and $Z$ to a common representation and (ii) decode $Y$ from this common representation. We evaluate our model on two tasks: (i) bridge transliteration and (ii) bridge captioning. We report promising results in both these applications and believe that this is a right step towards truly interlingua inspired encoder decoder architectures.


\end{abstract}

\section{Introduction}
Interlingua based MT \cite{inter1,inter2} relies on the principle that every language in the world can be mapped to a common linguistic representation. Further, given this representation, it should be possible to decode a target sentence in any language. This implies that given $n$ languages we just need $n$ decoders and $n$ encoders to translate between these $^nC_2$ language pairs. Note that even though we take inspiration from interlingua based MT, the focus of this work is not on MT. We believe that this idea is not just limited to translation but could be applicable to any kind of conversion involving multiple source and target languages and/or modalities (for example, transliteration, multilingual image captioning, multilingual image Question Answering, \textit{etc.}). Even though this idea has had limited success, it is still fascinating and considered by many as the holy grail of multilingual multimodal processing. 

It is interesting to consider the implications of this idea when viewed in the statistical context. For example, current state of the art statistical systems for MT \cite{pbsmt,hsmt,nmt-sota}, transliteration \cite{finch2015neural,shao2015boosting,nicolai2015multiple}, image captioning \cite{cap1,cap2}, \textit{etc.} require parallel data between the source and target views (where a view could be a language or some other modality like image). Thus, given $n$ views, we require $^nC_2$ parallel datasets to build systems which can convert from any source view to any target view. Obviously, this does not scale well in practice because it is hard to find parallel data between all $^nC_2$ views. For example, publicly available parallel datasets for transliteration \cite{Zhang:2012:RNM:2392777.2392779} cater to $<$ 20 languages. Similarly, publicly available image caption datasets are available only for English\footnote{\url{http://mscoco.org/dataset/$\#$download}} and German\footnote{\url{http://www.statmt.org/wmt16/multimodal-task.html}}.

This problem of resource scarcity could be alleviated if we could learn only $n$ statistical encoders and $n$ statistical decoders wherein (i) the encoded representation is common across languages and (ii) the decoders can decode from this common representation (akin to interlingua based conversion). As a small step in this direction, we consider a scaled down version of this generic $^nC_2$ conversion problem. Specifically, we consider the case where we have three views $X$, $Z$, $Y$ but parallel data is available only between $XZ$ and $ZY$ (instead of all $^3C_2$ parallel datasets). At test time, we are interested in generating natural language sequences in $Y$ starting from information available in $X$. We refer to this as the bridge setup as the language $Z$ here can be considered to be a bridge/pivot between $X$ and $Y$.  

An obvious solution to the above problem is to train a two-stage system which first converts from $X$ to $Z$ and then from $Z$ to $Y$. While this solution may work very well in practice (as our experiments indeed suggest) it is perhaps less elegant and becomes tedious as the number of views increases. For example, consider the case  of converting from $X$ to $Z$ to $R$ to $Y$. Instead, we suggest a neural network based model which simultaneously learns the following (i) a common representation for $X$ and $Z$ and (ii) decoding $Y$ from this common representation. In other words, instead of training two independent models using the datasets between $XZ$ and $ZY$, the model jointly learns from the two datasets. The resulting common representation learned for $X$ and $Z$ can be viewed as a vectorial analogue of the linguistic representation sought by interlingua based approaches. Of course, by no means do we suggest that this vectorial representation is a substitute for the rich linguistic representation but its easier to learn from parallel data (as opposed to a linguistic representation which requires hand crafted resources).       

Note that our work should not be confused with the recent work of \cite{cho-multi}, \cite{kevin} and \cite{Elliott}. The last two works in fact require 3-way parallel data between $X$, $Z$ and $Y$ and learn to decode sequences in $Y$ given both $X$ and $Z$. For example, at test time, \cite{Elliott} generate captions in German, given both (i) the image and (ii) its corresponding English caption. This is indeed very different from the problem addressed in this paper. Similarly, even though \cite{cho-multi} learn a single encoder per language and a single decoder per language they do not learn shared representations for multiple languages (only the attention mechanism is shared). Further, in all their experiments they require parallel data between the two languages of interest. Specifically, they do not consider the case of generating sentences in $Y$ given a sentence in $X$ when no parallel data is available between $X$ and $Y$.   

We present an empirical comparison of jointly trained models which explicitly aim for shared encoder representations with two-stage architectures. We consider two downstream applications (i) bridge transliteration and (ii) bridge caption generation. We use the standard NEWS 2012 dataset \cite{Zhang:2012:RNM:2392777.2392779} for transliteration. We consider transliteration between 12 languages pairs ($XY$) using English as the bridge ($Z$). Bridge caption generation is a new task introduced in this paper where the aim is to generate French captions for an image when no Image-French($XY$) parallel data is available for training. Instead training data is available between Image-English ($XZ$) and English-French ($ZY$). In both these tasks we report promising results. In fact, in our multilingual transliteration experiments we are able to beat the strong two-stage baseline in many cases. These results show potential for further research in interlingua inspired neural network architectures. We do acknowledge that a successful interlingua based statistical solution requiring only $n$ encoders and $n$ decoders is a much harder task whereas our work is only a small step in that direction.

\section{Related Work}

Encoder decoder based architectures for sequence to sequence generation were initially proposed in  \cite{enc-dec1,enc-dec2} in the context of Machine Translation (MT) and have also been successfully used for generating captions for images \cite{cap1}. However, such sequence to sequence models are often difficult to train as they aim to encode the entire source sequence using a fixed encoder representation. \newcite{attention} introduced attention based models wherein a different representation is fed to the decoder at each time step by focusing the attention on different parts of the input sequence. Such attention based models have been more successful than vanilla encoder-decoder models and have been used successfully for MT \cite{attention}, parsing \cite{parsing}, speech recognition \cite{speech}, image captioning \cite{cap2} among other applications. All the above mentioned works focus only on the case when there is one source and one target. The source can be image, text, or speech signal but the target is always a text sequence.

Encoder decoder models in a multi-source, single target setting have been explored by \cite{Elliott} and \cite{kevin}. Specifically, \newcite{Elliott} try to generate a German caption from an image and its corresponding English caption. Similarly, \newcite{kevin} focus on the problem of generating English translations given the same sentence in both French and German. We would like to highlight that both these models require three-way parallel data while we are focusing on situations where such data is not available. Single source, multi-target and multi-source, single target settings have been considered in \cite{multitasks2s}. Recent work by \newcite{cho-multi} explores multi-source to multi-target encoder decoder models in the context of Machine Translation. However, \newcite{cho-multi} focus on multi-task learning with a shared attention mechanism and the goal is to improve the MT performance for a pair of languages for which parallel data is available. This is clearly different from the goal of this paper which is to design encoder decoder models for a pair of languages for which no parallel data is available but data is available only between each of these languages and a bridge language.

Of course, in general the idea of pivot/bridge/interlingua based conversion is not new and has been used previously in several non-neural network settings. For example \cite{DBLP:conf/naacl/KhapraKB10} use a bridge language or pivot language to do machine transliteration. Similarly, \cite{pivot1,pivot2} do pivot based machine translation. Lastly, we would also like to mention the work in interlingua based Machine Translation \cite{inter1,inter2} which is clearly the inspiration for this work even though the focus of this work is not on MT.

The main theme explored in this paper is to learn a common representation for two views with the end goal of generating a target sequence in a third view. The idea of learning common representations for multiple views has been explored well in the past \cite{KlementievA2012,aps,HermannK2014,corrnet,DBLP:journals/corr/RajendranKCR15}. For example, \newcite{dcca} propose Deep CCA for learning a common representation for two views. \cite{aps,corrnet} propose correlational neural networks for common representation learning and \newcite{DBLP:journals/corr/RajendranKCR15} propose bridge correlational networks for multilingual multimodal representation learning. From the point of view of representation learning, the work of \newcite{DBLP:journals/corr/RajendranKCR15} is very similar to our work except that it focuses only on representation learning and does not consider the end goal of generating sequences in a target language.

\section{Models}
As mentioned earlier, one of the aims of this work is to compare a jointly trained model with a two stage model. We first briefly describe such a two stage encoder decoder architecture and then describe our model which is a correlation based jointly trained encoder decoder model.

\subsection{A two stage encoder-decoder model}

A two stage encoder-decoder is a straight-forward extension of sequence to sequence models \cite{enc-dec1,enc-dec2} to the bridge setup. Given parallel data between $XZ$ and $ZY$, a two stage model will learn a generative model for each of the pairs independently. For the purpose of this work, the source can be an image or text but the target is always a natural language text. For encoding an image, we simply take its feature representation obtained from one of the fully connected layers of a convolutional neural network and pass it through a feed-forward layer. On the other hand, for encoding the source text sequence, we use a recurrent neural network. The decoder is always a recurrent neural network which generates the text sequence, one token at a time.


\begin{figure}[h]
\begin{center}
\includegraphics[scale=0.6]{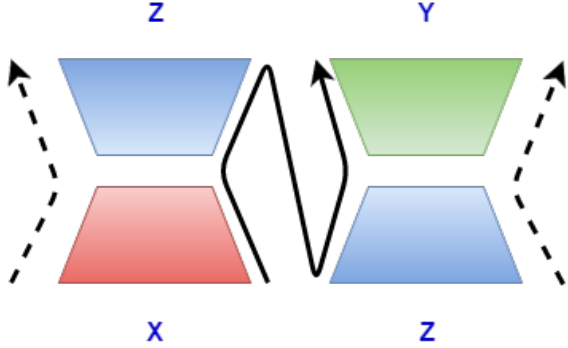}
\caption{Two stage encoder-decoder model. Dashed lines denote how the model is used during training time and solid line denotes the test time usage. We can see that two encoder-decoders are trained independently but used jointly during testing.}
\label{f1}
\end{center}

\end{figure}

Let the two training sets be $\mathcal{D}_1 = \{x_i, z_i\}_{i=1}^{N_1}$ and $\mathcal{D}_2 = \{z_i, y_i\}_{i=1}^{N_2}$ where $x_i \in X$, $y_i \in Y$ and $z_i \in Z$. Given $\mathcal{D}_1$, the first encoder learns to encode $x_i$ and decode the corresponding  $z_i$ from this encoded representation. The second encoder is trained independently of the first encoder and uses $\mathcal{D}_2$ to encode $z_i \in Z$ and decode the corresponding  $y_i \in Y$ from this encoded representation. These independent training processes are indicated by the dotted arrows in Figure \ref{f1}. 
At test time, the two stages are run sequentially. In other words, given $x_j$, we first encode it and decode $z_j$ from it using the first encoder-decoder model. This decoded $z_j$ is then fed to the second encoder-decoder model to generate $y_j$. This sequential test process is indicated by solid arrows in Figure \ref{f1}. 

While this two stage encoder-decoder model is a trivial extension of a single encoder-decoder model, it serves as a very strong baseline as we will see later in the experiments section.

\subsection{A correlation based joint encoder-decoder model}
While the above model works well in practice, it becomes cumbersome when more views are involved (for example, when converting from $U$ to $X$ to $Y$ to $Z$). We desire a more elegant solution which could scale even when more views are involved (although for the purpose of this work, we restrict ourselves to 3 views only). To this end, we propose a joint model which uses the parallel data $\mathcal{D}_1$ (as defined above) to learn one encoder each for $X$ and $Z$ such that the representations of $x_i$ and $z_i$ are correlated. In addition and simultaneously the model uses $\mathcal{D}_2$ and learns to decode $y_j$ from $z_j$. Note that this joint training has the advantage that the encoder for $Z$ benefits from instances in $\mathcal{D}_1$ and $\mathcal{D}_2$. 

Having given an intuitive explanation of the model, we now formally define the objective functions used during training. Given $\mathcal{D}_1 = \{x_i, z_i\}_{i=1}^{N_1}$, the model 
tries to maximize the correlation between the encoded representations of $x_i$ and $z_i$ as defined below. 

\begin{equation}
    \mathcal{J}_{\rm corr}(\theta) = - \lambda~{\rm corr}(s(h_X(X)),s(h_Z(Z)))
\end{equation}

where $h_X$ is the representation computed by the encoder for $X$ and $h_Z$ is the representation computed by the encoder for $Z$. As mentioned earlier, these encoders could be RNN encoders (in the case of text) and simple feedforward encoders (in the case of images). $s()$ is a standardization function which adjusts the hidden representations $h_X$ and $h_Y$ so that they have zero-mean and unit-variance. Further, $\lambda$ is a scaling hyper-parameter and $\rm corr$ is the correlation function as defined below:
\begin{equation}
\sum_{i=1}^N s(h_X(x_i))  s(h_Z(z_i))^T  
\end{equation}
We would like to emphasize that $s()$ ensures that the representations already have zero mean and unit variance and hence no separate standardization is required while computing the correlation.






\begin{figure}[h]
\begin{center}
\includegraphics[scale=0.6]{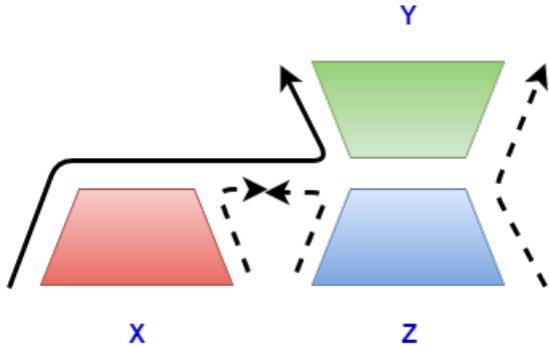}
\caption{Correlated encoder-decoder model. Dashed lines denote how the model is used during training time and solid line denotes the test time usage. We can see that during training, both the encoders are trained to produce correlated representations and the decoder for $Y$ is trained based on encoder $Z$. During test time only encoder for $X$ and decoder for $Y$ are used.}
\label{f2}
\end{center}

\end{figure} 

In addition to the above loss function, given $\mathcal{D}_2 = \{z_i, y_i\}_{i=1}^{N_2}$, the model minimizes the following cross entropy loss:

\begin{equation}
    \mathcal{J}_{\rm ce}(\theta) = \frac{1}{N_2} \sum_{k=1}^{N_2} P(y_k|z_k)
\label{eq:ce}    
\end{equation}

where

\begin{equation}
    P(y_k|z_k) = \prod_{i=1}^L P(y_{k_i}|y_{k_{<i}},z_k)
\end{equation}

where $L$ is the number of tokens in $y_k$.

The dotted lines in Figure \ref{f2} show the joint training process where the model simultaneously learns to compute correlated representations for $x_i$ and $z_i$ and decode $y_i$ from $z_i$. The testing process is shown by the solid lines wherein the model computes a hidden representation for $x_i$ and then decodes $y_i$ from it directly without transitioning through $z_i$.  

While training, we alternately pick mini-batches from $\mathcal{D}_1$ and $\mathcal{D}_2$ and use the corresponding objective function. Means and variances for the representations computed by the two encoders are updated at the end of every epoch based on the hidden representations of all instances in the training data. During the first epoch we assume the mean and variance to be 0 and 1. Note that $\lambda$ rescales the value of the correlation loss term so that it is in the same range as the value of the cross-entropy loss term. 



\section{Experiment 1: Bridge Transliteration}
\label{sec:BridgeTrans}

We consider the task of transliteration between two languages $X$ and $Y$ when no direct data is available between them but parallel data is available between $X$ \& $Z$ and $Z$ \& $Y$. In the following subsections we describe the datasets used for our task, the hyper-parameters considered for our experiments and results. 

\begin{table}[htbp]
\begin{center}
\footnotesize
\begin{tabular}{|c|c|c|c|}\hline
\textbf{Language Pair} & \textbf{Train-Set} & \textbf{Validation-Set} & \textbf{Test-Set} \\ \hline
\textbf{En-Hi} & 19918 & 500 & 1000 \\ \hline
\textbf{En-Ka} & 16556 & 500 & 1000 \\ \hline
\textbf{En-Ma} & 8500 & 500 & 1000 \\ \hline
\textbf{En-Ta} & 16857 & 500 & 1000 \\ \hline
\end{tabular}
\caption{Train, Validation and Test Splits of the NEWS 2012 Transliteration Corpus for the 4 Indian languages (\textbf{Hi}ndi, \textbf{Ka}nnada, \textbf{Ta}mil, \textbf{Ma}rathi)}
\label{Tab:NewsDataSplit}
\normalsize
\end{center}
\end{table}

\begin{table*}[htbp]
     \begin{center}
     \scriptsize
     \begin{tabular}{ccc} \\ 
     
    \begin{tabular}{|c|c|c|c|c|}
    \multicolumn{5}{c}{\textbf{Two Stage PBSMT}}\\ \hline
    \tiny{\backslashbox{src}{tgt}}& Hi & Ka & Ta & Ma \\ \hline
    Hi & \cellcolor{black!25} & 36.3 & 33.2 & 33.6 \\ \hline
    Ka & 41.3 & \cellcolor{black!25} & 32.1 & 26.2 \\ \hline
    Ta & 30.5 & 25.8 &  \cellcolor{black!25} & 19.2 \\ \hline
    Ma & 46.9 & 33.7 & 30.9 &  \cellcolor{black!25}\\ \hline
    \end{tabular}
    &
    \begin{tabular}{|c|c|c|c|c|}
    \multicolumn{5}{c}{\textbf{Two Stage Encoder Decoder}}\\ \hline
    \tiny{\backslashbox{src}{tgt}}& Hi & Ka & Ta & Ma \\ \hline
    Hi & \cellcolor{black!25} & \underline{42.1} & \textbf{\underline{43.4}} & \underline{34.8} \\ \hline
    Ka & \underline{46.2} & \cellcolor{black!25} & \textbf{\underline{42.9}} & \textbf{\underline{30.7}} \\ \hline
    Ta & \textbf{\underline{37.5}} & \textbf{\underline{34.8}} & \cellcolor{black!25} & \textbf{\underline{23.8}} \\ \hline
    Ma & \underline{45.8} & \underline{34.9} & \underline{31.7} & \cellcolor{black!25}\\ \hline
    \end{tabular}
    &
    \begin{tabular}{|c|c|c|c|c|}
    \multicolumn{5}{c}{\textbf{Correlational Encoder Decoder}}\\\hline
    \tiny{\backslashbox{src}{tgt}}& Hi & Ka & Ta & Ma \\ \hline
    Hi & \cellcolor{black!25} & \textbf{\underline{43.1}} & \underline{40.6} & \textbf{\underline{40.9}} \\ \hline
    Ka & \textbf{\underline{47.5}} & \cellcolor{black!25} & \underline{40.2} & \underline{27.9} \\ \hline
    Ta & \underline{33.6} & \underline{27.7} & \cellcolor{black!25} & 17.0 \\ \hline
    Ma & \textbf{\underline{59.0}} & \textbf{\underline{37.1}} & \textbf{\underline{34.5}} & \cellcolor{black!25}\\ \hline
    \end{tabular}
    
    \\
   \end{tabular}
   \caption{Transliteration Accuracy on the 12 language pairs involving (\textbf{Hi}ndi, \textbf{Ka}nnada, \textbf{Ta}mil, \textbf{Ma}rathi) for the three comparative methods (Two Stage PBSMT, Two Stage Encoder Decoder and the proposed Correlational Encoder Decoder model. An \underline{underlined} number in this table signifies that for that specific language pair the corresponding system is performing better than the Two Stage PBSMT model and the best performing system for any of the language-pairs is represented in \textbf{bold} font}
   \label{Tab:TranslitResults}
   \normalsize
   \end{center}
   
\end{table*} 

\subsection{Datasets}
We consider transliteration between 4 languages, \textit{viz.}, Hindi, Kannada, Tamil and Marathi resulting in $^4C_2 = 12$ language pairs. However, we do not use any direct parallel data between any of these languages. Instead we use the standard datasets available between English and each of these languages which were released as part of the NEWS 2012 shared task \cite{Zhang:2012:RNM:2392777.2392779}. Just to be explicit, for the task of transliterating from Hindi to Kannada, we construct $\mathcal{D}_1$ from the English-Hindi dataset and $\mathcal{D}_2$ from the English-Kannada dataset. Table \ref{Tab:NewsDataSplit} summarizes the sizes of these parallel datasets.  Fortunately, the English portion of the test set was common across all the 4 language pairs mentioned in Table \ref{Tab:NewsDataSplit}. This allowed us to easily create test sets for all the 12 language pairs. For example, if $h_i$ is the transliteration of the English word $e_i$ in the English-Hindi test set and $k_i$ is the transliteration of the same English word $e_i$ in the English-Kannada test set then we add $(h_i, k_i)$ as a transliteration pair in our Hindi-Kannada test set. In this way, we created test sets containing 1000 words for all the 12 language pairs. 

\subsection{Hyperparameters}
For the two stage encoder decoder model, we considered the following hyperparameters: embedding size $\in$ \{1024, 2048\} for characters, rnn hidden unit size $\in$ \{1024, 2048\}, initial learning rate $\in$ \{0.01, 0.001\} and batch size $\in$ \{32, 64\}. The numbers in bracket indicate the distinct values that we considered for each hyperparameter. Note that the embedding size and rnn size are always kept equal. All these parameters were tuned independently for the two stages using their respective validation sets. For the correlated encoder decoder model, in addition to the above hyperparamaters we also had $\lambda$ $\in$ [0.1, 1.0] as a hyperparameter. Here, we tuned the hyperparameters based on the performance on the validation set available between $ZY$ (since the correlated encoder decoder can also decode $y_i \in Y$ from $z_i \in Z$). Note that we do not use any parallel data between $XY$ for tuning the hyperparameters because the general assumption is that no parallel data is available between $XY$. 
We used Adam \cite{DBLP:journals/corr/KingmaB14} as the optimizer for all our experiments. 

\subsection{Results}
We compare our model with the following systems:

\noindent \textbf{1. Two Stage PBSMT}: Here, we train two PBSMT \cite{pbsmt} based transliteration systems using  $\mathcal{D}_1$ and $\mathcal{D}_2$. This is an additional baseline to see how well an encoder decoder architecture compares to a conventional PBSMT based system. We used Moses \cite{Koehn:2007:MOS:1557769.1557821} for building our PBSMT systems. The decoder parameters were tuned using the validation sets. Language model was trained on the target portion of the parallel corpus.

\noindent \textbf{2. Two Stage Encoder Decoder}: Here, we train two encoder decoder based transliteration systems using  $\mathcal{D}_1$ and $\mathcal{D}_2$ as described in Section \ref{sec:BridgeTrans}. 

Table \ref{Tab:TranslitResults} summarizes the accuracy (\% of correct transliterations) of the three systems in the bridge setup. We observe that in 6 out of the 12 language pairs our correlated model does better than the 2 stage encoder decoder model. Further, it does better than the two-stage PBSMT baseline in 11 out of the 12 language pairs. This is very encouraging especially because such 2-stage approaches are considered to be very strong baselines for these tasks \cite{DBLP:conf/naacl/KhapraKB10}. In general, the encoder decoder based approaches do better than PBSMT based systems. This is indeed the case even when we compare the performance of the PBSMT based system and the Encoder Decoder based system independently on the two stages (Table \ref{Tab:TranslitEnResults}). 

\begin{table}[htbp]
\begin{center}
\scriptsize
\begin{tabular}{|p{2.4cm}|p{1cm}|p{1cm}|} \hline
Source-Target Pair & \multicolumn{2}{|c|}{Accuracy(\%)}\\ \cline{2-3}
  & PBSMT & Encoder-Decoder \\ \hline
En-Hi & 51.7 & \textbf{61.6} \\ \hline
En-Ka & 45.3 & \textbf{53.7} \\ \hline
En-Ta & 50.0 & \textbf{57.7} \\ \hline 
En-Ma & 30.2 & \textbf{38.0} \\ \hline
Hi-En & 51.1 & \textbf{57.3} \\  \hline
Ka-En & 47.9 & \textbf{54.5} \\ \hline
Ta-En & 41.4 & \textbf{46.2} \\ \hline
Ma-En & \textbf{35.0} & 31.1 \\ \hline
\end{tabular}
\caption{Transliteration accuracy of the PBSMT system and the Encoder-Decoder model on the 4 Indian languages (\textbf{Hi}ndi, \textbf{Ka}nnada, \textbf{Ta}mil, \textbf{Ma}rathi) when transliterated from English and to English}
\normalsize
\label{Tab:TranslitEnResults}
\end{center}
\end{table}


\section{Experiment 2: Bridge Captioning}
We now introduce the task of bridge caption generation. The purpose of introducing this task is two-fold. Firstly, we feel that it is important to put things in perspective and demonstrate that while interlingua inspired encoder decoder architectures are a step in the right direction, much more work is needed when dealing with different modalities in a bridge setup. Secondly, we think that this is an important task which has not received any attention in the past. We would like to formally define and report some initial baselines to motivate further research in this area. The formal task definition is as follows: Generate captions for images in language $L_1$ (say, French) when no parallel data is available between images and $L_1$ but parallel data is available between Image-$L_2$ ($\mathcal{D}_1$) and between $L_1$-$L_2$ ($\mathcal{D}_2$) where $L_2$ is another language (say, English). 
In the following subsection we describe the datasets used for this task, the hyperparameters considered for our experiments and the results.

\begin{table*}[htbp]
\begin{center}
\scriptsize
\begin{tabular}{|l|c|c|c|c|c|c|}\hline
\textbf{Systems} & \textbf{BLEU-4} & \textbf{BLEU-3} & \textbf{BLEU-2} & \textbf{BLEU-1} & \textbf{ROUGE-L} & \textbf{CIDEr} \\ \hline
Pseudo Im-Fr  & 15.5 & 24.2 & 37.4 & 56.5 & 38.3 & 41.2 \\ \hline
Two Stage  & \textbf{16.6} & \textbf{25.7} & \textbf{39.0} & \textbf{58.3} & \textbf{39.5} & \textbf{49.1} \\ \hline
Correlational Encoder Decoder & 12.6 & 19.3 & 31.1 & 50.5 & 34.3 & 29.8  \\ \hline
\end{tabular}
\caption{Image Captioning performance in generating French caption for a given image for the three methods: Pseudo Im-Fr, Two Stage and our Correlational Encoder Decoder based model.}
\normalsize
\label{Tab:ImageCapResults}
\end{center}
\end{table*}


\subsection{Datasets}
Even though we do not have direct training data between Image-French, we need some test data to evaluate our model. For this, we use the Image-French test set recently released by \cite{DBLP:journals/corr/RajendranKCR15}. To create this data, they first merged the 80K images from the standard train split and 40K images from the standard valid split of MSCOCO data\footnote{\url{http://mscoco.org/dataset/$\#$download}}. They then randomly split the merged 120K images into train(118K), validation (1K) and test set (1K). They then collect French translations for all the 5 captions for each image in the test set using crowdsourcing. CrowdFlower\footnote{\url{https://make.crowdflower.com}} was used as the crowdsourcing platform and they solicited one French and one German translation for each of the 5000 captions using native speakers. Note that \cite{DBLP:journals/corr/RajendranKCR15} report results for cross modal search and do not address the problem of crosslingual image captioning.

In our model, for $D_1$ we use the same train(118K), validation (1K) and test sets (1K) as defined in \cite{DBLP:journals/corr/RajendranKCR15} and explained above. Choosing $\mathcal{D}_2$ was a bit more tricky. Initially we considered the corpus released as part of WMT'12 \cite{callisonburch-etal_WMT:2012} which contains roughly 44M English-French parallel sentences from various sources including News, parliamentary proceedings, etc. However, our initial small scale experiments showed that this does not work well because there is a clear mismatch between the vocabulary of this corpus and the vocabulary that we need for generating captions. Also the vocabulary is much larger (at least an order higher than what we need for image captioning) and it thus hampers training. Further, the average length and structure of these sentences is also very different from captions. Domain shift in MT is itself a challenging problem (not to mention the added complexity in a multimodal bridge setup). It was unrealistic to expect our model to work in the presence of these orthogonal complexities. 

To isolate these issues and evaluate our model in a controlled environment, we needed a parallel corpus which had very similar characteristics to that observed in captions. Since we did not have such a corpus at our disposal we decided to follow \cite{DBLP:journals/corr/RajendranKCR15} and use a pseudo parallel corpus between English-French. Specifically, we take the English captions from the MSCOCO data and translate them to French using the publicly available translation system provided by IBM\footnote{\url{http://www.ibm.com/smarterplanet/us/en/ibmwatson}}. Note that our model still does not see direct parallel data between Image and French during training. We acknowledge that this is not the ideal thing to do but it is good enough to do a proof-of-concept evaluation of our model and understand its potential. We, of course, account for the liberty taken here by comparing with equally strong baselines as discussed later in the results section.  

\subsection{Hyperparameters}
Our model has the following hyperparameters: embedding size, batch size, hidden representation size, $\lambda$ and learning rate. Based on experiments involving direct Image-to-English caption generation we observed that the following parameters work well : {embedding size = 512,  batch size = 80, rnn hidden unit size = 512, and learning rate = 4e-4 with Adam \cite{DBLP:journals/corr/KingmaB14} as the optimizer. We just retained these hyper-parameters and did not tune them again for the bridge setup. We tuned the value of $\lambda$ by evaluating the correlation loss on the Image-English validation set. Again, we do not use any Image-French data for tuning any hyperparameters.

\begin{table*}[htbp]
\scriptsize
    \begin{center}
    \begin{tabular}{|p{0.5cm}|p{2.2cm}|p{2.2cm}|p{2.2cm}|p{2.2cm}|p{2.2cm}|p{2.2cm}|} \hline
    \tiny{\textbf{\rotatebox[origin=l]{90}{MSCOCO Images}}} & \includegraphics[width=2.2cm,height=2.2cm]{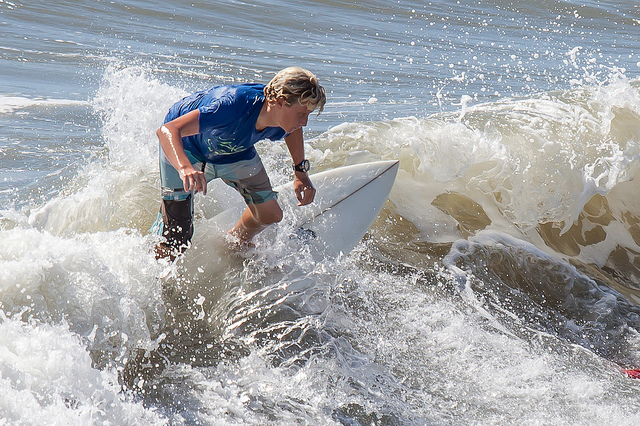} &
    \includegraphics[width=2.2cm,height=2.2cm]{"images/mscocoimages/COCOtrain2014000000063036"} &
    \includegraphics[width=2.2cm,height=2.2cm]{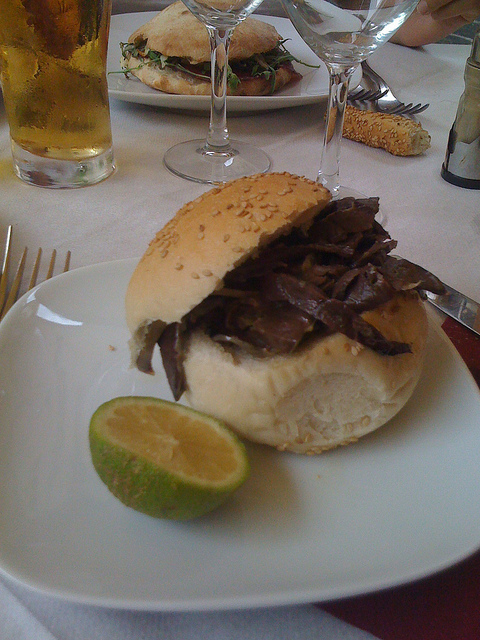} &
    \includegraphics[width=2.2cm,height=2.2cm]{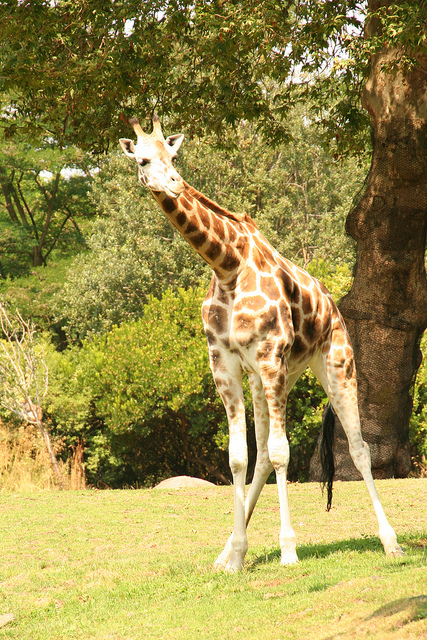} &
    \includegraphics[width=2.2cm,height=2.2cm]{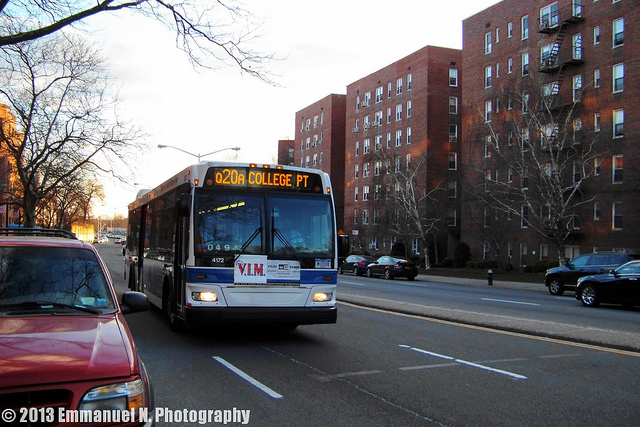} &
    \includegraphics[width=2.2cm,height=2.2cm]{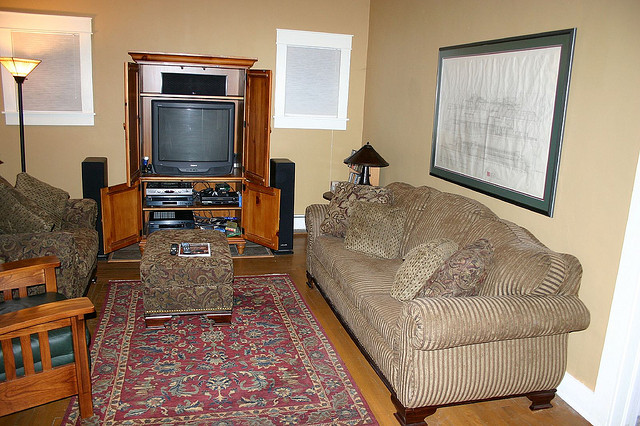}\\ \hline
    
 \tiny{ \rotatebox[origin=c]{90}{\textbf{\parbox{1.8cm}{Correlational\\ Encoder Decoder}}}}  & Un homme est surfer sur une vague dans l'océan & Un skateur est en train de décoller sur un skateboard & Une plaque avec un sandwich et un verre de bière & Une girafe est debout dans la poussière près d'une arborescence & Un bus de transport en commun dans une rue de la ville & Un salon avec un canapé , une table et un téléviseur  \\ \hline
   \tiny{ \textbf{\rotatebox[origin=c]{90}{\parbox{1.3cm}{Two-Stage}}}} & Un homme circonscription une vague sur une planche de surf  & Un homme circonscription un skateboard sur une rampe en bois &  Une plaque de nourriture sur une table avec un verre de vin & Une girafe debout dans un champ avec des arbres en arrièreplan & Un bus double sandwich au volant dune rue & Un salon avec un canapé fauteuil et une télévision \\ \hline
  \tiny{ \textbf{ \rotatebox[origin=c]{90}{\parbox{1.7cm}{Pseudo Im-Fr}}}} & Un internaute dans une combinaison isothermique est circonscription une vague &  Un jeune garçon circonscription un skateboard dans un parc & Une plaque de nourriture sur une table en bois & Une girafe debout à côté d'un autre girafe dans une zone & Un bus ville faire baisser une rue de la ville & Un salon avec un canapé , une table et un canapé \\ \hline
    
    \end{tabular}
    \caption{Example captions generated by the three methods on a sample set of MSCOCO test images}
    \label{Tab:ImageCapExamples}
    \end{center}
    \normalsize
    
\end{table*}    

\subsection{Results}
We now present the results of our experiments where we compare with the following strong baselines.

\noindent \textbf{1. Two Stage} : Here we use a Show \& Tell model \cite{cap1} trained using $D_1$ to generate an English caption for the image. We then translate this caption into French using IBM's translation system as described above.

\noindent \textbf{2. Pseudo Im-Fr} : Here we train an Image-to-French Show \& Tell model \cite{cap1} by pairing the images in the MSCOCO dataset with their pseudo French captions generated by translating the English captions into French (using IBM's translation system).

We observe that our model is unable to beat the two strong baselines described above but still comes close to their performance. We believe this reinforces our belief in this line of research and hopefully more powerful models (perhaps attention based) could eventually surpass these two baselines.

As a qualitative evaluation of our model, Table \ref{Tab:ImageCapExamples} shows the captions generated by our model. It is exciting that even in a complex multimodal bridge setup the model is able to capture correlations between Images and English sentences and further decode relevant French captions from a given image.  




\section{Conclusion}
In this paper, we considered the problem of pivot based sequence generation. Specifically, we are interested in generating sequences in a target language starting from information in a source view. However, no direct training data is available between the source and target views but training data is available between each of these views and a pivot view. To this end, we take inspiration from interlingua based MT and propose a neural network based model which explicitly maximizes the correlation between the source and pivot view and simultaneously learns to decode target sequences from this correlated representation. We evaluate our model on the task of bridge transliteration and show that it outperforms a strong two-stage baseline for many language pairs. Finally, we introduce the task of bridge caption generation and report promising initial results. We hope this new task will fuel further research in this area. 

As future work, we would like to go beyond simple encoder decoder based correlational models. For example, we would like to apply the idea of correlation to attention based encoder decoder models. The ideas expressed here can also be applied to other tasks such as bridge translation, bridge Image QA, etc. However, for these tasks, additional issues such as larger vocabulary sizes, complex sentence structures, non-monotonic alignments between source and target language pairs need to be addressed. The model proposed here is just a beginning and much more work is needed to cater to these complex tasks.   




\bibliography{emnlp2016}

\begin{thebibliography}{}

\bibitem[\protect\citename{Andrew \bgroup et al.\egroup }2013]{dcca}
Galen Andrew, Raman Arora, Jeff Bilmes, and Karen Livescu.
\newblock 2013.
\newblock Deep canonical correlation analysis.
\newblock {\em ICML}.

\bibitem[\protect\citename{Bahdanau \bgroup et al.\egroup }2014]{attention}
Dzmitry Bahdanau, Kyunghyun Cho, and Yoshua Bengio.
\newblock 2014.
\newblock Neural machine translation by jointly learning to align and
  translate.
\newblock {\em CoRR}, abs/1409.0473.

\bibitem[\protect\citename{Callison-Burch \bgroup et al.\egroup
  }2012]{callisonburch-etal_WMT:2012}
Chris Callison-Burch, Philipp Koehn, Christof Monz, Matt Post, Radu Soricut,
  and Lucia Specia.
\newblock 2012.
\newblock Findings of the 2012 workshop on statistical machine translation.
\newblock In {\em Seventh Workshop on Statistical Machine Translation}, WMT,
  pages 10--51, Montr{\'e}al, Canada.

\bibitem[\protect\citename{Chandar \bgroup et al.\egroup }2014]{aps}
Sarath Chandar, Stanislas Lauly, Hugo Larochelle, Mitesh~M Khapra, Balaraman
  Ravindran, Vikas Raykar, and Amrita Saha.
\newblock 2014.
\newblock An autoencoder approach to learning bilingual word representations.
\newblock In {\em Proceedings of NIPS}.

\bibitem[\protect\citename{Chandar \bgroup et al.\egroup }2016]{corrnet}
Sarath Chandar, Mitesh~M. Khapra, Hugo Larochelle, and Balaraman Ravindran.
\newblock 2016.
\newblock Correlational neural networks.
\newblock {\em Neural Computation}, 28(2):257 -- 285.

\bibitem[\protect\citename{Chiang}2005]{hsmt}
David Chiang.
\newblock 2005.
\newblock A hierarchical phrase-based model for statistical machine
  translation.
\newblock In {\em {ACL} 2005, 43rd Annual Meeting of the Association for
  Computational Linguistics, Proceedings of the Conference, 25-30 June 2005,
  University of Michigan, {USA}}.

\bibitem[\protect\citename{Cho \bgroup et al.\egroup }2014]{enc-dec1}
Kyunghyun Cho, Bart van Merrienboer, {\c{C}}aglar G{\"{u}}l{\c{c}}ehre, Dzmitry
  Bahdanau, Fethi Bougares, Holger Schwenk, and Yoshua Bengio.
\newblock 2014.
\newblock Learning phrase representations using {RNN} encoder-decoder for
  statistical machine translation.
\newblock In {\em Proceedings of the 2014 Conference on Empirical Methods in
  Natural Language Processing, {EMNLP} 2014, October 25-29, 2014, Doha, Qatar,
  {A} meeting of SIGDAT, a Special Interest Group of the {ACL}}, pages
  1724--1734.

\bibitem[\protect\citename{Chorowski \bgroup et al.\egroup }2015]{speech}
Jan Chorowski, Dzmitry Bahdanau, Dmitriy Serdyuk, Kyunghyun Cho, and Yoshua
  Bengio.
\newblock 2015.
\newblock Attention-based models for speech recognition.
\newblock In {\em Advances in Neural Information Processing Systems 28: Annual
  Conference on Neural Information Processing Systems 2015, December 7-12,
  2015, Montreal, Quebec, Canada}, pages 577--585.

\bibitem[\protect\citename{Dorr \bgroup et al.\egroup }2010]{inter2}
Bonnie~J. Dorr, Rebecca~J. Passonneau, David Farwell, Rebecca Green, Nizar
  Habash, Stephen Helmreich, Eduard~H. Hovy, Lori~S. Levin, Keith~J. Miller,
  Teruko Mitamura, Owen Rambow, and Advaith Siddharthan.
\newblock 2010.
\newblock Interlingual annotation of parallel text corpora: a new framework for
  annotation and evaluation.
\newblock {\em Natural Language Engineering}, 16(3):197--243.

\bibitem[\protect\citename{Elliott \bgroup et al.\egroup }2015]{Elliott}
Desmond Elliott, Stella Frank, and Eva Hasler.
\newblock 2015.
\newblock Multi-language image description with neural sequence models.
\newblock {\em CoRR}, abs/1510.04709.

\bibitem[\protect\citename{Finch \bgroup et al.\egroup }2015]{finch2015neural}
Andrew Finch, Lemao Liu, Xiaolin Wang, and Eiichiro Sumita.
\newblock 2015.
\newblock Neural network transduction models in transliteration generation.
\newblock {\em Proceedings of NEWS 2015 The Fifth Named Entities Workshop},
  page~61.

\bibitem[\protect\citename{Firat \bgroup et al.\egroup }2016]{cho-multi}
Orhan Firat, KyungHyun Cho, and Yoshua Bengio.
\newblock 2016.
\newblock Multi-way, multilingual neural machine translation with a shared
  attention mechanism.
\newblock {\em CoRR}, abs/1601.01073.

\bibitem[\protect\citename{Hermann and Blunsom}2014]{HermannK2014}
Karl~Moritz Hermann and Phil Blunsom.
\newblock 2014.
\newblock Multilingual models for compositional distributed semantics.
\newblock In {\em Proceedings of the 52nd Annual Meeting of the Association for
  Computational Linguistics, {ACL} 2014, June 22-27, 2014, Baltimore, MD, USA,
  Volume 1: Long Papers}, pages 58--68.

\bibitem[\protect\citename{Khapra \bgroup et al.\egroup
  }2010]{DBLP:conf/naacl/KhapraKB10}
Mitesh~M. Khapra, A.~Kumaran, and Pushpak Bhattacharyya.
\newblock 2010.
\newblock Everybody loves a rich cousin: An empirical study of transliteration
  through bridge languages.
\newblock In {\em Human Language Technologies: Conference of the North American
  Chapter of the Association of Computational Linguistics, Proceedings, June
  2-4, 2010, Los Angeles, California, {USA}}, pages 420--428.

\bibitem[\protect\citename{Kingma and Ba}2014]{DBLP:journals/corr/KingmaB14}
Diederik~P. Kingma and Jimmy Ba.
\newblock 2014.
\newblock Adam: {A} method for stochastic optimization.
\newblock {\em CoRR}, abs/1412.6980.

\bibitem[\protect\citename{Klementiev \bgroup et al.\egroup
  }2012]{KlementievA2012}
Alexandre Klementiev, Ivan Titov, and Binod Bhattarai.
\newblock 2012.
\newblock {Inducing Crosslingual Distributed Representations of Words}.
\newblock In {\em Proceedings of the International Conference on Computational
  Linguistics (COLING)}.

\bibitem[\protect\citename{Koehn \bgroup et al.\egroup }2003]{pbsmt}
Philipp Koehn, Franz~Josef Och, and Daniel Marcu.
\newblock 2003.
\newblock Statistical phrase-based translation.
\newblock In {\em {HLT-NAACL}}.

\bibitem[\protect\citename{Koehn \bgroup et al.\egroup
  }2007]{Koehn:2007:MOS:1557769.1557821}
Philipp Koehn, Hieu Hoang, Alexandra Birch, Chris Callison-Burch, Marcello
  Federico, Nicola Bertoldi, Brooke Cowan, Wade Shen, Christine Moran, Richard
  Zens, Chris Dyer, Ond\v{r}ej Bojar, Alexandra Constantin, and Evan Herbst.
\newblock 2007.
\newblock Moses: Open source toolkit for statistical machine translation.
\newblock In {\em Proceedings of the 45th Annual Meeting of the ACL on
  Interactive Poster and Demonstration Sessions}, ACL '07, pages 177--180,
  Stroudsburg, PA, USA. Association for Computational Linguistics.

\bibitem[\protect\citename{Luong \bgroup et al.\egroup }2015a]{multitasks2s}
Minh{-}Thang Luong, Quoc~V. Le, Ilya Sutskever, Oriol Vinyals, and Lukasz
  Kaiser.
\newblock 2015a.
\newblock Multi-task sequence to sequence learning.
\newblock {\em CoRR}, abs/1511.06114.

\bibitem[\protect\citename{Luong \bgroup et al.\egroup }2015b]{nmt-sota}
Minh{-}Thang Luong, Hieu Pham, and Christopher~D. Manning.
\newblock 2015b.
\newblock Effective approaches to attention-based neural machine translation.
\newblock {\em CoRR}, abs/1508.04025.

\bibitem[\protect\citename{Nicolai \bgroup et al.\egroup
  }2015]{nicolai2015multiple}
Garrett Nicolai, Bradley Hauer, Mohammad Salameh, Adam St~Arnaud, Ying Xu, Lei
  Yao, and Grzegorz Kondrak.
\newblock 2015.
\newblock Multiple system combination for transliteration.
\newblock {\em Proceedings of NEWS 2015 The Fifth Named Entities Workshop},
  page~72.

\bibitem[\protect\citename{Nirenburg}1994]{inter1}
Sergei Nirenburg.
\newblock 1994.
\newblock Pangloss: {A} machine translation project.
\newblock In {\em Human Language Technology, Proceedings of a Workshop held at
  Plainsboro, New Jerey, USA, March 8-11, 1994}.

\bibitem[\protect\citename{Rajendran \bgroup et al.\egroup
  }2015]{DBLP:journals/corr/RajendranKCR15}
Janarthanan Rajendran, Mitesh~M. Khapra, Sarath Chandar, and Balaraman
  Ravindran.
\newblock 2015.
\newblock Bridge correlational neural networks for multilingual multimodal
  representation learning.
\newblock {\em CoRR}, abs/1510.03519.

\bibitem[\protect\citename{Shao \bgroup et al.\egroup }2015]{shao2015boosting}
Yan Shao, J{\"o}rg Tiedemann, and Joakim Nivre.
\newblock 2015.
\newblock Boosting english-chinese machine transliteration via high quality
  alignment and multilingual resources.
\newblock {\em Proceedings of NEWS 2015 The Fifth Named Entities Workshop},
  page~56.

\bibitem[\protect\citename{Sutskever \bgroup et al.\egroup }2014]{enc-dec2}
Ilya Sutskever, Oriol Vinyals, and Quoc~V. Le.
\newblock 2014.
\newblock Sequence to sequence learning with neural networks.
\newblock In {\em Advances in Neural Information Processing Systems 27: Annual
  Conference on Neural Information Processing Systems 2014, December 8-13 2014,
  Montreal, Quebec, Canada}, pages 3104--3112.

\bibitem[\protect\citename{Vinyals \bgroup et al.\egroup }2015a]{parsing}
Oriol Vinyals, Lukasz Kaiser, Terry Koo, Slav Petrov, Ilya Sutskever, and
  Geoffrey~E. Hinton.
\newblock 2015a.
\newblock Grammar as a foreign language.
\newblock In {\em Advances in Neural Information Processing Systems 28: Annual
  Conference on Neural Information Processing Systems 2015, December 7-12,
  2015, Montreal, Quebec, Canada}, pages 2773--2781.

\bibitem[\protect\citename{Vinyals \bgroup et al.\egroup }2015b]{cap1}
Oriol Vinyals, Alexander Toshev, Samy Bengio, and Dumitru Erhan.
\newblock 2015b.
\newblock Show and tell: {A} neural image caption generator.
\newblock In {\em {IEEE} Conference on Computer Vision and Pattern Recognition,
  {CVPR} 2015, Boston, MA, USA, June 7-12, 2015}, pages 3156--3164.

\bibitem[\protect\citename{Wu and Wang}2007]{pivot1}
Hua Wu and Haifeng Wang.
\newblock 2007.
\newblock Pivot language approach for phrase-based statistical machine
  translation.
\newblock In {\em {ACL} 2007, Proceedings of the 45th Annual Meeting of the
  Association for Computational Linguistics, June 23-30, 2007, Prague, Czech
  Republic}.

\bibitem[\protect\citename{Xu \bgroup et al.\egroup }2015]{cap2}
Kelvin Xu, Jimmy Ba, Ryan Kiros, Kyunghyun Cho, Aaron~C. Courville, Ruslan
  Salakhutdinov, Richard~S. Zemel, and Yoshua Bengio.
\newblock 2015.
\newblock Show, attend and tell: Neural image caption generation with visual
  attention.
\newblock In {\em Proceedings of the 32nd International Conference on Machine
  Learning, {ICML} 2015, Lille, France, 6-11 July 2015}, pages 2048--2057.

\bibitem[\protect\citename{Zhang \bgroup et al.\egroup
  }2012]{Zhang:2012:RNM:2392777.2392779}
Min Zhang, Haizhou Li, A.~Kumaran, and Ming Liu.
\newblock 2012.
\newblock Report of news 2012 machine transliteration shared task.
\newblock In {\em Proceedings of the 4th Named Entity Workshop}, NEWS '12,
  pages 10--20, Stroudsburg, PA, USA. Association for Computational
  Linguistics.

\bibitem[\protect\citename{Zhu \bgroup et al.\egroup }2014]{pivot2}
Xiaoning Zhu, Zhongjun He, Hua Wu, Conghui Zhu, Haifeng Wang, and Tiejun Zhao.
\newblock 2014.
\newblock Improving pivot-based statistical machine translation by pivoting the
  co-occurrence count of phrase pairs.
\newblock In {\em Proceedings of the 2014 Conference on Empirical Methods in
  Natural Language Processing, {EMNLP} 2014, October 25-29, 2014, Doha, Qatar,
  {A} meeting of SIGDAT, a Special Interest Group of the {ACL}}, pages
  1665--1675.

\bibitem[\protect\citename{Zoph and Knight}2016]{kevin}
Barret Zoph and Kevin Knight.
\newblock 2016.
\newblock Multi-source neural translation.
\newblock {\em CoRR}, abs/1601.00710.

\end{thebibliography}
\bibliographystyle{emnlp2016}

\end{document}